# Stereo Vision Based Single-Shot 6D Object Pose Estimation for Bin-Picking by a Robot Manipulator

Yoshihiro Nakano

*Abstract*—We propose a fast and accurate method of 6D object pose estimation for bin-picking of mechanical parts by a robot manipulator. We extend the single-shot approach to stereo vision by application of attention architecture. Our convolutional neural network model regresses to object locations and rotations from either a left image or a right image without depth information. Then, a stereo feature matching module, designated as Stereo Grid Attention, generates stereo grid matching maps. The important point of our method is only to calculate disparity of the objects found by the attention from stereo images, instead of calculating a point cloud over the entire image. The disparity value is then used to calculate the depth to the objects by the principle of triangulation. Our method also achieves a rapid processing speed of pose estimation by the single-shot architecture and it is possible to process a 1024 × 1024 pixels image in 75 milliseconds on the Jetson AGX Xavier implemented with half-float model. Weakly textured mechanical parts are used to exemplify the method. First, we create original synthetic datasets for training and evaluating of the proposed model. This dataset is created by capturing and rendering numerous 3D models of several types of mechanical parts in virtual space. Finally, we use a robotic manipulator with an electromagnetic gripper to pick up the mechanical parts in a cluttered state to verify the validity of our method in an actual scene. When a raw stereo image is used by the proposed method from our stereo camera to detect black steel screws, stainless screws, and DC motor parts, i.e., cases, rotor cores and commutator caps, the bin-picking tasks are successful with 76.3%, 64.0%, 50.5%, 89.1% and 64.2% probability, respectively.

## I. INTRODUCTION

A robot vision system for 6D object pose estimation is a key role to expand applications of industrial robot manipulators. In particular, bin-picking is a robotic task of picking objects up in a cluttered state and used to automate many industrial operations. Moreover, many industrial parts have weakly textured surface, and, therefore, depth images, point cloud data measured by a depth camera, are typically used to calculate their position and orientation [1, 2]. However, such measurement equipments limit the wide application in factory sites by higher cost. Furthermore, as described in an earlier report [1], measuring dense point clouds or depth images on glossy objects is difficult using optical scanners that are commonly used as industrial depth cameras. Therefore, 3D depth cameras are not universally used for detecting all objects in industry.

For low-cost applications, a monocular camera based methods has been reported for 6D object pose estimation. There are a lot of research on typical methods such as

Yoshihiro Nakano is with Tokyo Research and Development Center, MinebeaMitsumi Inc., Tokyo 108-8330, Japan (email: ynakano.wm@minebeamitsumi.com).

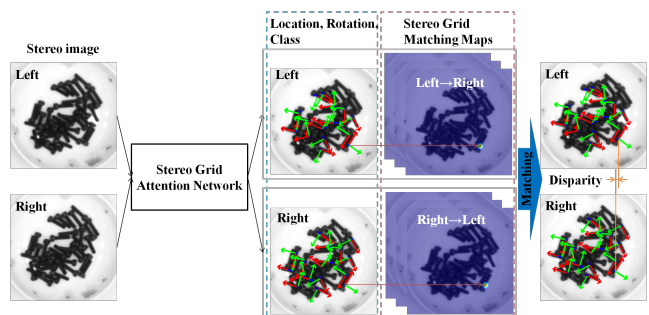

Figure 1. Overview of the proposed method for object 6D pose estimation. Stereo grid attention network, the proposed CNN model, outputs objects locations on a 2D image, rotation, and classification scores to each left and right image. Additionally, the CNN generates stereo grid matching maps. Then, disparity is calculated by subtracting the estimated horizontal locations in the matched grid cells on the left and the right images. Finally, the depth is calculated by disparity and stereo camera parameters.

feature-based method [3, 4] and template-based method [5, 6]. However, they have shortcomings for detecting weakly textured objects and occlusions. In recent years, deep learning has been more fascinating for object pose estimation [7, 8, 9, 10, 11] than the preceding methods. The deep learning methods are robust against occlusion, less textured object surface, and changes in surrounding environments such as lighting conditions. However, when using monocular cameras, the depth estimation error becomes worse as the object smaller, whereas robot vision requires more accurate locations of smaller objects in bin-picking tasks.

On the other hand, stereo vision can keep the estimation error being constant even if the object is so small. Recently, the Region Proposal Network (RPN) [12] is extended to a stereo vision system for 3D object detection to support autonomous driving [13]. A particularly interesting aspect of the method is that it can detect accurately 3D objects directly based on stereo vision without generating a depth image. However, RPN-based iterative computation model [12] is generally slower than single-shot detection models [14, 15, 16].

In this paper, we propose a passive stereo vision based object pose estimation method that is less expensive than the typical industrial 3D camera based methods. Our method can detect low-textured, glossy and low-reflective objects, and obtain more accurate depth measurement than the monocular camera based methods. We extend the concept of single-shot object detection to stereo vision for fast 6D object pose estimation. Figure 1 presents an overview of this stereo matching and pose estimation method. Our convolutional neural network (CNN) model, which we call the Stereo Grid Attention Network, classifies object types and predicts the object locations and orientations, from a pair of stereo images, i.e., a left image and a right image. Then, the object depth values are only calculated from disparity between regressed locations in the pair images. An attention-based architecture is used to calculate a matching score to find the same objects between the pair images, and depth information is calculated only for the centroid of the objects with a higher attention score instead of the entire image. The computational load of our method is consequently much lighter than the conventional ones. (See section III for more details). Furthermore, inspired by success with a synthetic dataset [17] for training a pose estimation model [10], we attempt to train our model only with a synthetic dataset created using a 3D object model in virtual space. A robot vision system with our method can then be applied to detecting real industrial objects that has a 3D model used for the training. More specifically, we create synthetic datasets of M4 black screws, stainless screws and three types of motor parts using Unreal Engine 4, a 3D game engine, for training of the CNN model. Our plug-in into Unreal Engine has a simple randomizing function and the ability to synchronize two cameras perfectly to create a dataset for stereo applications. As a particular advantage of the application, only a simple background color of the dataset image must be considered for the industrial scene. Results of the object detection in the real images and bin-picking by a robot manipulator are presented in sections IV. They show the possibility of training the CNN model in a sufficiently robust manner by randomizing the color and material parameters. In this way, training datasets can be created by using simple settings for many industrial parts.

## II. RELATED WORKS

### A. Monocular-Image-Based Object Pose Estimation

Conventionally, the feature-based methods [3, 4] and the template-based methods [5, 6] have been studied. In recent years, deep learning based methods are applied to object pose estimation, and the estimate is robust against occlusion, weakly textured object surface, and changes in light and background. Among those models, PoseCNN [11] and BB8 [8] can be divided into to two steps, segmentation and pose estimation. Segmentation is to detect each object in an image, whereas pose estimation predicts a pose of the detected objects. Moreover, single-shot object detection approaches [14, 15] are extended to object pose estimation tasks [7, 9, 10]. Some earlier reports [9, 10] have proposed computationally lighter models with the same prediction accuracy. Some studies have been conducted to predict coordinates of corners on a projected 3D bounding box [8, 9, 10] and to restore a object location and orientation in 3D with the PnP algorithm. It can train an accurate and general model that is independent of the internal parameters of a specific camera, but a post-processing treatment is required after the prediction. Some earlier reports also have described [7, 11] a method of direct prediction of the object pose without the post-processing treatment.

### B. Deep-Learning-Based Stereo Matching

To extend the CNN model to pose estimation with stereo images, we briefly review its application to stereo cameras. In the field of stereo vision depth sensor, a typical subject is to generate a dense disparity map. A CNN design is to shift and stack one feature map and concatenate it to the other feature map [18, 19]. Then, disparities are predicted from the maps with the CNN, however, this method is computationally heavy when calculating a disparity value between far apart pixels. On the one hand, image-patch-based stereo matching technique has also been proposed [20, 21]. However, such iterative image-patch matching between left and right images takes computation time as well. A study of stereo-based super-resolution [22] recently proposed an efficient stereo matching architecture, parallax-attention module (PAM) based on the attention mechanism [23, 24, 25].

### C. Stereo-Vision-Based Object Detection

In the field of autonomous driving, a method has been proposed for direct detection of 3D objects based on stereo images [13] without expensive light detection and ranging (LiDAR) equipment. A 3D object detection method usually consists of depth map creation and a 3D object detection from the map, and the two steps of calculation are integrated with this method into a single step. Furthermore, the performance of the detection algorithm is free from constraints of LiDAR capacity and usual stereo vision algorithms.

## III. APPROACH

We propose a CNN model that can estimate accurate object poses with efficient stereo matching processes. The model outputs the classes, stereo grid matching maps, and poses without depth. Disparity calculation methods are such that the estimated location of an object in the right image is subtracted from the estimated one in the left image. The proposed approach finds the pairs of estimated locations of the same objects between the left and the right images with the stereo grid matching maps. For one earlier study [13], left-right anchor pairs were regressed by RPN. However, RPN-based iterative processing is slower than the single-shot approach. Our method can find the pair between far apart locations in the pair images by an efficient attention-architecture [23, 24, 25]. For one study [22], super-resolution images are reconstructed from stereo images based on the attention architecture. However, computation loads are heavy for whole image stereo matching. Our method enables sub-pixel order stereo matching using sparse feature maps and object pose estimation simultaneously. Figure 2 portrays our CNN model, the Stereo Grid Attention Network, with its three stages: a feature extraction stage, a stereo feature matching stage, and a prediction stage. Finally, post processing is done for depth calculation.

### A. Feature Extraction Stage

At this stage, separating the features of the object and the surrounding objects must be done while preserving the object features of location and orientation. To extract global surroundings information and local location information, CNN for feature map extraction include down-sampling, up-sampling and skip-connection architecture. Only the last layer in extracted feature maps is adopted for the next stage. Industrial bin-picking applications do not require the use of feature pyramids as in typical object detection models [14, 16] to detect objects because the size of the object projected on the image is roughly known. Compared to the adoption of feature pyramids, the fixed size feature map according to the target object size can reduce the computational load. Furthermore, by setting the grid cell size as smaller than the object size, false detections can be suppressed because difficult situations arising from multiple objects in the same grid cell are reduced.

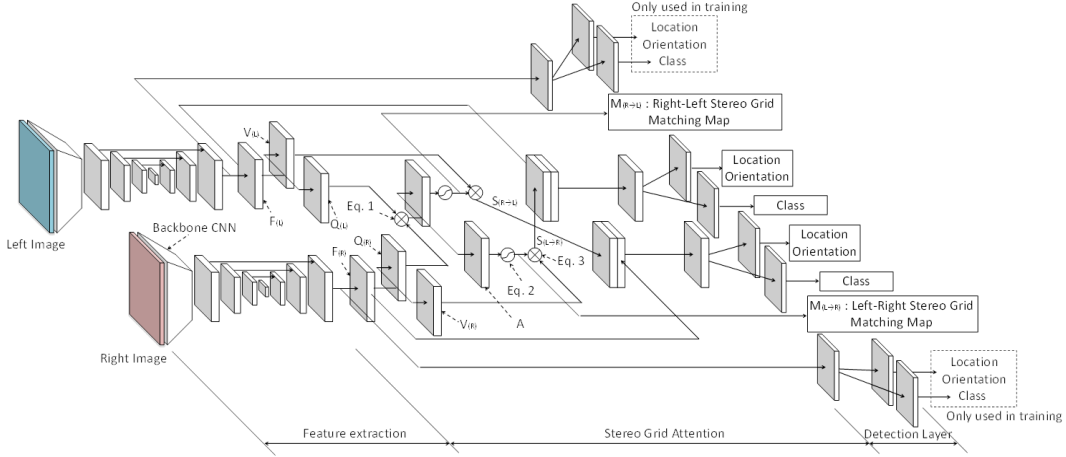

Figure 2. The architecture of Stereo Grid Attention Network. The model includes the three stages. In the first stage, feature maps are extracted by the CNN module, which has down-sampling, up-sampling, and skip-connection. In the second stage, the Stereo Grid Attention module creates the stereo grid matching maps, and concatenate the left and the right features. In the third stage, the detection layer predicts objects locations, rotations, and classification scores.

*B. Stereo Grid Attention Stage*

First, a CNN is used to generate two feature maps **Q** and **V** from output of the feature extraction stage **F**. extracted in ach of left and right images. Feature maps $\mathbf{Q}_{(L)}$ and $\mathbf{Q}_{(R)}$ are used for stereo matching between the left and right images, and $\mathbf{V}_{(L)}$ and $\mathbf{V}_{(R)}$ for feature connection between the two; the subscripts L and R denote the left and right images, respectively. Here, Q is a matrix in three dimensions, i.e., $Q \in \mathbf{R}^{H \times W \times C}$, and $\mathbf{Q}_h \in \mathbf{R}^{W \times C}$ is an *h*th array extracted from **Q** in the first dimension. A correlation matrix $\mathbf{A}_h$ is then defined between the left and right feature maps as

$$\mathbf{A}_h = \mathbf{Q}_{(L)h} \mathbf{Q}_{(R)h}^{\mathrm{T}}. \quad (1)$$

In the left side of Eq. (1), the matrix elements represent a correlation of the left to the right feature maps in the row direction, and a correlation of the right to the left in the column direction. The correlation $\mathbf{A}_h$ is normalized by the softmax function to give a stereo grid matching map of the left to the right feature maps as

$$\mathbf{M}_{(L \to R)hw} = \frac{[\exp(\mathbf{A}_{hw0}) \ \exp(\mathbf{A}_{hw1}) \ \cdots \ \exp(\mathbf{A}_{hwW-1})]}{\sum_{i=0}^{W-1} \exp(\mathbf{A}_{hwi})}, \quad (2)$$

where, $\mathbf{M}_{(L \to R)hw}$ is a vector representing the matching score between the left grid cell of height index *h* and width index *w* and the right grid cells of height index *h*. Consequently, $\mathbf{M}_{(L \to R)}$, the left-right stereo grid matching map, includes all the left to right stereo grid matching scores. A stereo grid matching map of the right to the left feature maps, $\mathbf{M}_{(R \to L)}$, is obtained by applying the softmax function in the row of transposed $\mathbf{A}_h$. Furthermore, $\mathbf{V}_{(R)h}$ is multiplied by $\mathbf{M}_{(L \to R)h}$ to yield

$$\mathbf{S}_{(L \to R)h} = \mathbf{M}_{(L \to R)h} \mathbf{V}_{(R)h}, \quad (3)$$

where the features of the right image that are matched with the features of each grid cell of the left image are strongly reflected to $\mathbf{S}_{(L \to R)h}$. Finally, $\mathbf{S}_{(L \to R)}$ is concatenated to the left feature map $\mathbf{F}_{(L)}$. The same concatenation process is calculated from the right feature map to the left feature map.

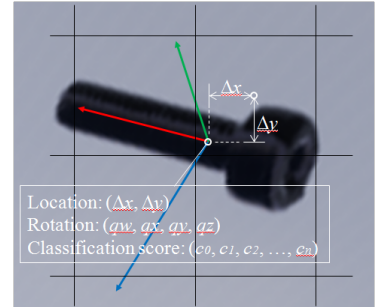

Figure 3. Definition of grid-cell outputs. Each grid cell has outputs of location, rotation, and classification score. The location is the objects position offsets from the center of the grid cell. The rotation, regressed with quaternion, is the orientation of local coordinate system from the reference coordinate system. The classification score is the probability that centroid each type of object exist on each grid cell.

*C. Detection Stage*

In this detection stage, the network layers predict the object pose and classes from the features created in sections III.*A* and III.*B*. Figure 3 shows grid-cell network outputs, that is, classification scores ($c_0$, $c_1$, ..., $c_n$), object locations ($\Delta x$, $\Delta y$) as an offset from the cell center, and object rotation expressed by quaternions ($qw$, $qx$, $qy$, $qz$). The quaternions are then defined as rotation relative to a reference frame; the *z*-axis is aligned with the line of sight from the camera center to the object center (Figure 4). The classification score $c_0$ means that there is no target object found in a cell, and the scores $c_1$, ..., $c_n$ target object types. Note that each grid cell detects only one object to predict matching between the right and the left outputs using the stereo grid matching maps.

*D. Depth Calculation*

Object depth values are calculated using the stereo grid matching maps and the estimated object locations calculated in the prior stage. Algorithm 1 present a calculation procedure to reduce mismatching of object locations between the stereo images. All the steps in the algorithm are calculated to yield a disparity map.

**Algorithm 1** Disparity Calculation

**input:**

$C_{(L)}$, $C_{(R)}$: classification score maps of a left and a right image, defined in section III.C

$M_{(L \to R)}$, $M_{(R \to L)}$: a left-right and a right-left stereo grid matching maps, defined in section III.B

$X_{(L)}$, $X_{(R)}$: horizontal location maps on a left and a right image, reconstructed by locations $\Delta x$ defined in section III.C

$t$: threshold of classification score to object detection

**output:**

$D$: disparity map

**procedure**

   **for** $h \in [0, 1, \ldots, H\text{-}1]$ **do**

      **for** $w \in [0, 1, \ldots, W\text{-}1]$ **do**

         $c_{(L)} = \text{argmax}_i(C_{(L)hwi})$, $s_{(L)} = \max_i(C_{(L)hwi})$

         **if** $c_{(L)}\ != 0$ **and** $s_{(L)} > t$ **then**

            $r = \text{argmax}_i(M_{(L \to R)hwi})$

            $c_{(R)} = \text{argmax}_i(C_{(R)hri})$, $s_{(R)} = \max_i(C_{(R)hri})$

            **if** $c_{(R)} == c_{(L)}$ **and** $s_{(R)} > t$ **then**

                $w' = \text{argmax}_i(M_{(R \to L)hri})$

                **if** $w == w'$ **then**

                    $D_{hw} = X_{(L)hw} - X_{(R)hr}$

                **endif**

            **endif**

         **endif**

      **endfor**

   **endfor**

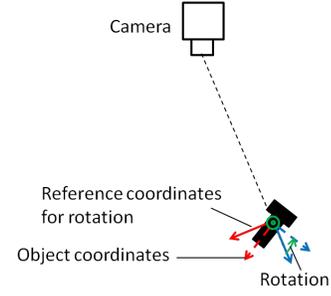

Figure 4. Definition of the reference coordinates system for the object rotation. The coordinates system show the *z*-axis as a straight line draw from the camera center to the object center.

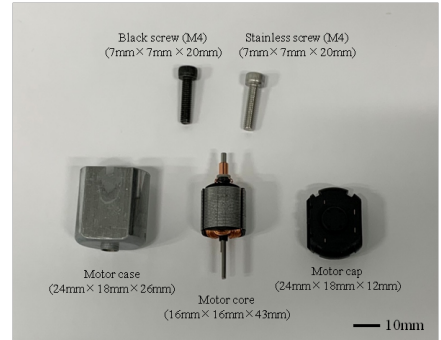

Figure 5. The target objects for experiment. The objects include the black screw, stainless screw and three types of DC motor parts.

Finally, the depth is calculated from disparity map **D** using the principle of triangulation. In this method, the disparity is given by the subtraction between the regressed object locations, and the resultant stereo matching is naturally with a sub-pixel order of precision. Our method uses sparse stereo matching to give a light computation load.

## IV. EXPERIMENT

Figure 5 shows the target objects as black steel screws, stainless screws and DC motor parts. The target objects include low textured, glossy, and low reflective ones. They are a few centimeter or smaller in length and, and it is necessary to estimate an object location with a millimeter order accuracy to pick them up. We train our proposed model using a dataset created on Unreal Engine 4 using 3D models of the target objects. Next, the object depth values are estimated from real images by the trained model and the estimation accuracy is statistically evaluated with a laser instrument. Finally, we operated a task of bin-picking by a robot manipulator using the proposed model.

### A. Hardware setting

Table I shows specifications of a stereo camera used in the experiments. We train the proposed model for object pose estimation in a depth range from 600 mm to 900 mm. In this stereo camera setup, a disparity error of one pixel becomes a 1.7 mm depth error at a distance of 600 mm, and 3.8mm at 900 mm.

TABLE I. SPECIFICATIONS OF THE STEREO CAMERA

| Image format | Number of pixels | Field of view | Baseline |
|---|---|---|---|
| Grayscale | 1280×960 | 33.4° × 25.4° | 100mm |

### B. Implementation Details

For this work, we adopt the U-Net [26] architecture for the feature extraction CNN. We use VGG[27]-like model as a backbone network with a batch normalization layer [28] and a squeeze-and-excitation block [29]. The original VGG network is designed for a color image input and modified for a grayscale image to reduce number of inputs. Additionally, channels of some layers are reduced for calculation load reduction from the original VGG. In the training term, a 32 × 32 × 256 feature map is extracted from a 512 × 512 image in the feature extraction stage. In the stereo grid attention stage, the feature maps for matching, $Q_{(L)}$ and $Q_{(R)}$, are implemented in a size of 32 × 32 × 64, and the feature maps for concatenating, $V_{(L)}$ and $V_{(R)}$, are implemented in a size of 32 × 32 × 128. Because the target depth range is determined, the outside features of the expected matching range is masked when calculating the stereo grid matching maps to reduce the influence of the extra features. The detection layers contain two classes to detect the screws, $c_0$: no objects found in a grid cell and $c_1$: screw, and four classes to detect the motor parts, $c_0$: no objects in a cell, $c_1$: motor case, $c_2$: rotor core and $c_3$: commutator cap. In the evaluation term, width and height of the input image and all feature maps are doubled from the training ones. Namely, the input image size is 1024 × 1024 for the proposed model in the evaluation. The blank pixels in the vertical direction that causes by the lack of number of pixels in the raw image, and in the horizontal direction that causes by the clipping with offset of described in IV.*D* are expanded to 1024 × 1024 by applying the zero padding.

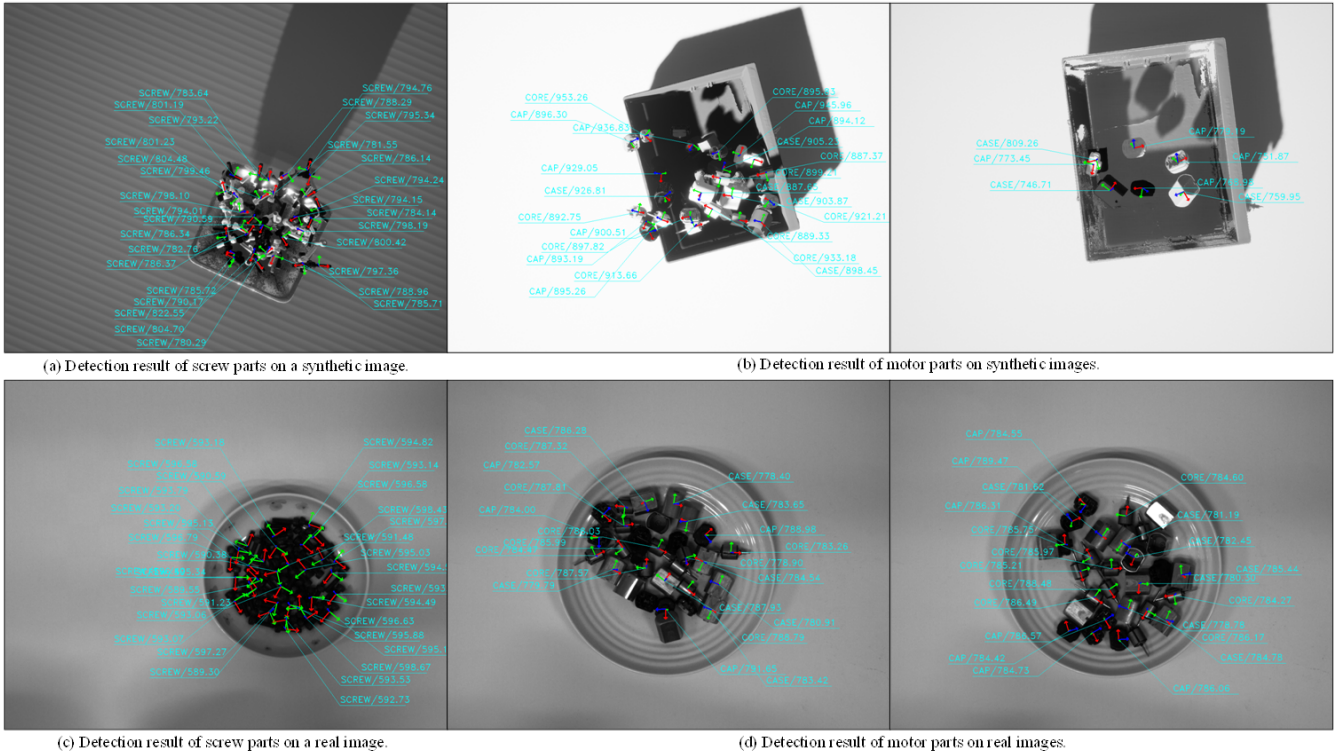

Figure 6. Examples of predicted result by the evaluation model rendered on real and synthetic images of the left of stereo pairs.

### C. Training Datasets

3D models of the target objects and their containers are imported into Unreal Engine 4 to create our datasets. To use the stereo camera described in section IV.*A*, set the stereo camera with the same parameters in Unreal Engine 4. In a virtual environment, the objects are spawned in a random number under the stereo camera, and the container moves randomly on a floor. The objects, the container and the floor are set to a random variable both in a grayscale and in a material texture. The stereo cameras also randomly move together and the capture timing are synchronized with each other. A light moves randomly as well, rotating the orientation and changing the intensity. The dataset is created to include captured images and ground truth data of object location and rotation and of classes as a text file. The screw and the motor core are assumed to be rotationally symmetric, and the rotation is fixed at zero around the symmetric axis. It is assume that an object is detectable if the object has a size of area beyond a threshold number of pixels in an image. The threshold is inversely proportional to a square of the depth of the object and defined as 1200, 3600, 2400, and 2400 pixels, respectively, for the screws, the motor case, the motor core, and the commutator cap at 500mm depth.

### D. Train

The proposed model is trained using the dataset presented in section IV.*C*. The input images for the proposed model is cropped to a size of 512 × 512 pixels at random locations. To make a pair of stereo images, a left image is cropped and offset by 355 pixels, equivalent to disparity at 600mm depth, to the right side from that of the right image. Similarly to other single-shot object detection models [14, 16], the output values of the location and rotation are divided by fixed values, and the ideas of hard negative mining are applied to train our model. For this work, we set the values of divisor to 0.1 for location and 0.2 for rotation. We also use the ratio of 1:1 between the negatives and the positives in hard negative mining. Unlike other single-shot approaches, the intersection-over-union-based anchor box matching is not performed. Instead, the grid cell closest to the centroid of the screw is selected to output the target object class. A smooth L1 loss function is used for regressing location and rotation, and a softmax cross-entropy loss for the class. To strengthen the connection between the right and the left features in the stereo grid attention, loss functions are evaluated only on an object appearing both in the stereo images. The backbone network has been trained with ImageNet 2012 dataset [30], and our model is then trained by transfer learning.

### E. Evaluation

The evaluation model processes stereo images of 1024 × 1024 pixels using the weight parameters trained in section IV.*D*. The model is implemented on the Jetson AGX Xavier (NVIDIA Corp.) by TensorRT in FP16 mode of operation, and the processing time is 75 milliseconds. The object location and orientation are evaluated by the trained model, and the results are designated by an arrow on the synthetic and the real images (Figure 6). It is shown that an object pose for real images can be predicted by the model trained only with the synthetic images.

To measure the estimation error of depth, we randomly place objects on a flat surface and capture their images. They do not overlap with one another to align the height of all of them. Figure 7 shows root-mean-square (RMS) error of depth and disparity for objects with the classification score greater than 0.6, and it is also shown that the estimation errors are less than one pixel in disparity, a few millimeters in depth, for most of types of objects in a range from 600 to 900 mm.

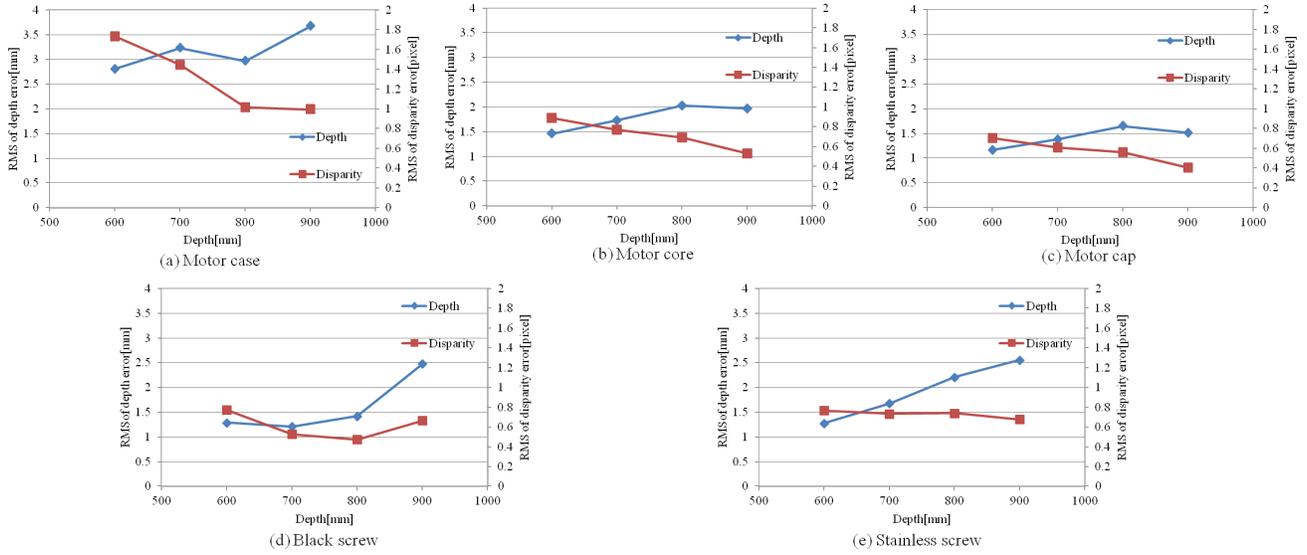

Figure 7. The root mean square (RMS) of depth error and disparity error.

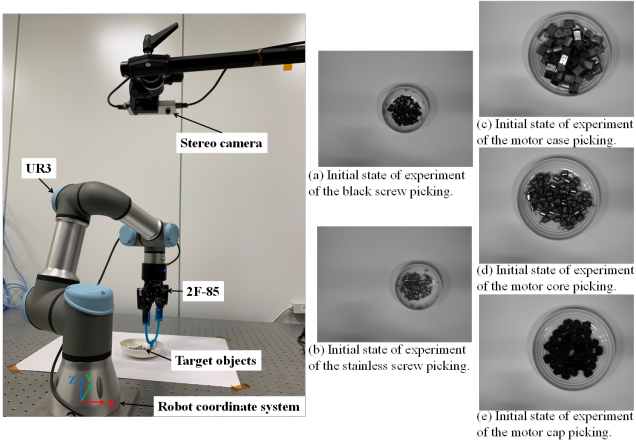

Figure 8. Setup for the bin-picking task by a robot manipulator and initial states of the target objects for the bin-picking trials.

### F. Bin-Picking by a Robot Manipulator

Apart from the estimation accuracy, it is the most important to pick up the target objects using a robot manipulator. In order to verify the effectiveness of the proposed model, we show an actual robot can picked up the target objects with the network outputs. Left side of figure 8 presents the hardware settings. Robot manipulator UR3 (Universal Robots, Corp.) is used for the task of pick-up, and gripper 2F-85 (Robotiq) is attached to the tip of the manipulator. A stereo camera is set at the height of 700mm above a table surface. Target objects are cluttered and placed on a dish as a container just beneath the camera. Trial starts with 50 objects on the dish and counts successes and failures until all objects are picked up. Initial states of target objects for the trials are shown on the right side of figure 8. The trial results of bin-picking task are presented in Table II. In a failure case, it is judged as "Deeper" when the gripper finger touches the dish only slightly even if the object is grasped by a gripper. It is expressed as "Shallower" when the estimated depth is too shallow to reach the target objects. "Others" include the cases where the surroundings block the finger from the target and where the rotation error causes a grasping failure. The motor case and the commutator cap have a cuboid shape and are more difficult to pick up than the cylindrical objects, a screw and a rotor core. Target screws are more difficult to pick up than the rotor core because of simply smaller. To summarize the results, the bin-picking tasks are successful with a probability of 76.3 %, 64.0 %, 50.5 %, 89.1 %, and 64.3 %, respectively, for the black screws, the stainless screws, the motor cases, the motor cores, and the commutator caps. It is verified that the proposed method can detect and pick-up low textured, glossy, and low reflective objects with millimeter-order depth accuracy.

TABLE II. THE SUCCESSFUL PROBABILITY OF THE OBJECT BIN-PICKING TASKS.

| Object Type | Results of Bin-Picking | | | |
|---|---|---|---|---|
| | *Success* | *Deeper* | *Shallower* | *Others* |
| Black screws | 76.3% | 8.5% | 11.8% | 3.4% |
| Stainless screws | 64.0% | 15.7% | 12.5% | 7.8% |
| Motor case | 50.5% | 0.0% | 19.6% | 29.9% |
| Motor core | 89.1% | 1.8% | 3.6% | 5.5% |
| Motor cap | 64.2% | 3.1% | 3.1% | 29.6% |

### V. CONCLUSION

We propose a new method of object pose estimation in 6D via stereo cameras, and the method is based on the single-shot approach with an attention architecture. Our method can accurately estimate the location of weakly textured mechanical parts cluttered with large quantities. The evaluation results show that our model is trained solely by synthetic data generated in the virtual space, and that it also provides a good estimate of object pose with real images.

This method is readily usable in factories because training data can be generated for mechanical parts with weak textures in virtual space. Our method also works well with low-cost stereo cameras, and is expected to useful for expanding the field of activity of robot manipulators.